\documentclass{article}

\usepackage{arxiv}

\usepackage[utf8]{inputenc} 
\usepackage[T1]{fontenc}   
\usepackage{hyperref}      
\usepackage{url}            
\usepackage{booktabs}       
\usepackage{amsfonts}       
\usepackage{nicefrac}       
\usepackage{microtype}     
\usepackage{lipsum}
\usepackage{graphicx}
\usepackage{subfigure}
\usepackage{multirow}

\title{BERTomelo: Your Portuguese Encoder Best Friend}

\author{
 Rennê Ruan Alves Oliveira \\
  Department of Computer Science\\
  University of Brasília\\
  Brasília - DF, Brazil\\
  \texttt{renne.oliveira@aluno.unb.br} \\
   \And
 Gustavo Cordeiro Galvão Van Erven \\
  Department of Computer Science\\
  University of Brasília\\
  Brasília - DF, Brazil\\
  \texttt{gustavo.erven@cgu.gov.br} \\
  \And
 Luís Paulo Faina Garcia \\
  Department of Computer Science\\
  University of Brasília\\
  Brasília - DF, Brazil\\
  \texttt{luis.garcia@unb.br} \\
}

\begin{document}
\maketitle
\begin{abstract}
Encoders have become the state of the art for multiple NLP tasks, especially those requiring deep contextual understanding. While multilingual models offer broad coverage, dedicated monolingual encoders are essential for capturing the unique lexical and syntactic nuances of specific languages. For Portuguese, however, existing monolingual options like BERTimbau and Albertina have not kept pace with recent architectural breakthroughs, often lagging behind English benchmarks in scalability and efficiency. This work introduces BERTomelo, a next-generation monolingual encoder pre-trained from scratch and specifically optimized for the Portuguese language. By leveraging the ModernBERT architecture, BERTomelo overcomes the limitations of previous models, offering Base and Large versions with a 1,024-token context window and hardware-level optimizations like FlashAttention and alternating attention mechanisms. The model was trained on ClassiCC-PT, a massive, high-quality Portuguese corpus of 106 million documents, ensuring superior alignment with the language's contemporary usage. The results demonstrate that BERTomelo not only outperforms previous Portuguese encoders but also provides a more robust and efficient alternative to massive multilingual models in downstream tasks such as STS and NER.
\end{abstract}

\keywords{BERT \and Portuguese BERT \and ModernBERT \and Encoders.}

\section{Introduction}
\label{sec:intro}

The transformer architecture plays a central role in Natural Language Processing (NLP)~\cite{Vaswani2017}. Originally designed as a sequence-to-sequence model for machine translation, it consists of two main components: an encoder, which produces a representation of the input, and a decoder, which utilizes the encoder's output and the preceding tokens to generate a new sequence. Through subsequent architectural advancements, the encoder and decoder have frequently been decoupled to serve as standalone models. Decoders constitute the backbone of generative models, whereas encoders are primarily used for text representation in tasks such as classification and information retrieval~\cite{Minaee2025}. 

Encoders remain a cornerstone of Artificial Intelligence, particularly in supporting generative frameworks through Retrieval-Augmented Generation. These architectures are frequently fine-tuned to excel within specialized knowledge domains or utilized as feature extractors to produce deep, contextualized embeddings for clustering and similarity analysis. Significantly more compact than their decoder counterparts, encoder models offer superior versatility, feasible fine-tuning, and accelerated inference speeds~\cite{Antoun2025}.

Among the most influential architectures in this context, the Bidirectional Encoder Representations from Transformers (BERT)~\cite{Devlin2018} serves as a foundational architecture for the development of encoder models. It is pre-trained using Masked Language Modeling (MLM), predicting tokens replaced by a special \texttt{[MASK]} token, and Next Sentence Prediction (NSP), which determines if one sentence logically follows another. While the original BERT was trained on an English corpus, subsequent multilingual versions, such as mBERT, were released to enhance versatility. However, while multilingual models offer a broad linguistic range and overall performance, monolingual models pre-trained on a corpus dedicated to a single language demonstrate superior performance on tasks within that language~\cite{Rust2020,Conneau2019}. 

Since the release of BERT, several encoder models for Portuguese have been proposed, including BERTimbau~\cite{Souza2020}, BERTugues~\cite{MazzaZago2024}, and PeLLE~\cite{deMello2024}. However, most of these models adhere to the original BERT architecture, typically using standard attention with limited 512-token context windows and constrained corpora. Although approaches such as Albertina~\cite{Rodrigues2023} and DeBERTinha~\cite{Campiotti2023} introduce improvements in the pre-training, gaps in training efficiency and context size remain open issues. In contrast, the state-of-the-art in English advances more rapidly, with models such as ModernBERT~\cite{Warner2024} incorporating advanced architectures and training strategies. It integrates mechanisms absent from existing Portuguese models, such as FlashAttention~\cite{Dao2022}, Alternating Attention~\cite{Warner2024}, and Unpadding~\cite{Zeng2022}. These components are combined to provide a model with an 8,192-token context size window. It also utilizes Rotary Positional Embeddings (RoPE), which allow for extrapolation to unseen token positions during inference.

To address this gap, BERTomelo is introduced as a new suite of Portuguese BERT-based encoders that integrates several features from the recent ModernBERT architecture~\cite{Warner2024}. It utilizes this architecture as a foundation for training from scratch on the Classified Common Crawl Corpus for Portuguese (ClassiCC-PT)~\cite{Almeida2025}, a large dataset composed of curated and filtered texts. Two variants are proposed: a Base version, consisting of 22 layers, 12 attention heads, and a hidden size of 768; and a Large model, featuring 28 layers, 16 attention heads, and a hidden size of 1,024. In contrast to previous approaches, BERTomelo is presented with a 1,024-token context window for both versions. Experimental results indicate that both models achieve state-of-the-art performance in Named Entity Recognition (NER) and Semantic Textual Similarity (STS) tasks, as well as competitive metrics in Recognizing Textual Entailment (RTE) tasks compared to other Portuguese and multilingual models.

The remainder of the paper is structured as follows: Section \ref{sec:relworks} reviews related work on encoder models; Section \ref{sec:bertomelo} presents the characteristics of the BERTomelo models; Section \ref{sec:experiments} describes the methodology used to train and evaluate BERTomelo models; Section \ref{sec:results} reports both pre-training and downstream task results, comparing BERTomelo with other models; and finally, Section \ref{sec:conclusion} presents the conclusions and directions for future work.

\section{Related Works}
\label{sec:relworks}

The introduction of the Transformer architecture represented a paradigm shift in NLP~\cite{Vaswani2017}. Designed to overcome the sequential bottlenecks of recurrent neural networks, the Transformer enables the processing of entire sequences through parallelized matrix operations. By leveraging the attention mechanism, in which each token attends to all others within a sequence, this architecture eliminates temporal dependencies, facilitates highly scalable training, and enables the model to capture complex contextual and syntactic dependencies across the data.

Built upon the Transformer's encoder block, the BERT model~\cite{Devlin2018} leverages bidirectional representations to capture comprehensive contextual information from both preceding and succeeding tokens simultaneously. The architecture consists of a stack of encoder blocks. To handle sequence order, it employs absolute positional embeddings. BERT's pre-training is driven by two key objectives: MLM and NSP, utilizing specialized tokens such as \texttt{[MASK]}, \texttt{[CLS]}, and \texttt{[SEP]}. The architecture was originally released in two primary configurations: BERT Base, which comprises 12 layers, a hidden size of 768, and 12 attention heads, totaling 110 million parameters; and BERT Large, which features 24 layers, a hidden size of 1,024, and 16 attention heads, for a total of 340 million parameters. Both versions support a 512-token context window. While initially trained on an English corpus, the framework was later extended to support a broad of languages with the release of mBERT~\cite{Devlin2018}.

BERT established itself as the pioneer for subsequent BERT-family models. Later architectures sought to address its limitations, introducing more robust approaches and improving scalability. One major evolution was the Robustly Optimized BERT Approach (RoBERTa)~\cite{Liu2019} model, motivated by the premise that BERT is significantly under-trained. RoBERTa introduces a dynamic masking strategy and removes the NSP task, but its most significant contributions are scaling pre-training, optimizing training hyperparameters, and increasing the training batch size from 256 to 8,192. The RoBERTa model highlighted the need for a robust pretraining protocol, demonstrating that superior performance can be achieved by scaling the model's training regime. Initially, RoBERTa was trained on an English corpus, but the approach was later applied to multilingual datasets to create XLM-RoBERTa~\cite{Conneau2019}.

Following RoBERTa, the Decoding-enhanced BERT with Disentangled Attention (DeBERTa)~\cite{He2023} model introduced fundamental modifications to the attention mechanism by altering how positional information is processed. DeBERTa introduced Disentangled Attention, which treats a token's content and its relative position as separate vectors. In this framework, the attention weight between two tokens also depends on their relative distance. The architecture evolved through three major iterations: V1 established these core concepts; V2 expanded the vocabulary to 128,000 tokens via SentencePiece~\cite{Kudo2018} and scaled to 1.5 billion parameters; V3 replaced the MLM objective with the Replaced Token Detection objective.

Focusing on computational efficiency, the MosaicBERT~\cite{Portes2024} model integrated architectural advances to optimize pre-training, achieving accuracy metrics comparable to the RoBERTa model with drastically reduced costs. Key modifications included replacing absolute positional embeddings with Attention With Linear Biases~\cite{Press2022}, which apply distance-proportional penalties to attention scores to enable context window extrapolation. Additionally, MosaicBERT pioneered the use of FlashAttention~\cite{Dao2022} and Unpadding~\cite{Zeng2022}. These structural changes placed MosaicBERT on a new efficiency-performance Pareto frontier.

Expanding upon these concepts, ModernBERT~\cite{Warner2024} represents a generational shift in encoder-only architectures, utilizing a 2 trillion token pre-training corpus and extending the 8,192-token context window. A ``deep and narrow'' hardware-aware design optimizes inference speed and parameter efficiency, while the integration of RoPE~\cite{Su2023} enables sequence length extrapolation. The model employs Alternating Attention alongside efficiency-focused techniques such as Unpadding~\cite{Zeng2022} with Sequence Packing~\cite{Krell2021} and FlashAttention~\cite{Dao2022}, effectively reducing memory bottlenecks and computational waste. ModernBERT outperforms DeBERTa-V3~\cite{He2023} on the GLUE benchmark and processes short contexts twice as fast. The architecture is available in two versions: the Base model, with 22 layers, a hidden size of 768, and 12 attention heads, comprising approximately 149 million parameters; and the Large model, with 28 layers, a hidden size of 1,024, and 16 attention heads, totaling roughly 395 million parameters.

Beyond architectural innovations, other works have focused on the language representation of BERT-based models. Multilingual models are founded on the hypothesis of cross-lingual transfer, where patterns learned from high-resource languages could benefit under-represented ones. However, as the number of supported languages grew, a fundamental capacity trade-off emerged. This phenomenon, termed the ``Curse of Multilinguality''~\cite{Conneau2019}, occurs when the inclusion of diverse languages degrades per-language performance relative to models trained on dedicated language datasets. The superiority of monolingual approaches has been empirically corroborated by specialized models such as FinBERT~\cite{Virtanen2019}, CamemBERT~\cite{Martin2020}, BETO~\cite{Canete2023}, and AraBERT~\cite{Antoun2021}. For Brazilian Portuguese, this efficacy was most notably demonstrated by BERTimbau~\cite{Souza2020}, which remains a benchmark in the field.

Trained on the BrWaC~\cite{Filho2018} corpus with a custom-built vocabulary, BERTimbau~\cite{Souza2020} served as the primary baseline for Portuguese BERT models. It was released in two versions: the Base version, with 12 layers, a hidden size of 768, and 12 attention heads, with approximately 110 million parameters, initialized from mBERT~\cite{Devlin2018}; and the Large version, with 24 layers, a hidden size of 1,024, and 16 attention heads, with approximately 335 million parameters, which uses English BERT~\cite{Devlin2018} weights as a starting point. The Base version was trained for 1 million steps with a 512-token context window, while the Large version was trained for 900,000 steps at a 128-token length before expanding to a 512-token context window for the final 100,000 steps. Despite the limitations of a 512-token context window and a pre-training corpus of only 2.6 billion tokens, BERTimbau continues to be widely used for Portuguese applications. Its direct evolution, BERTuguês~\cite{MazzaZago2024}, optimized the tokenization process by eliminating rare characters and natively integrating emojis. Following the architectural leap of the DeBERTa model, Albertina~\cite{Rodrigues2023} emerged. Featuring approximately 900 million parameters, the largest for Brazilian Portuguese, it redefined the state-of-the-art by outperforming BERTimbau in five out of six evaluated downstream tasks.

Despite these advancements, BERT-based models for Brazilian Portuguese still lag behind their counterparts in other languages, particularly in terms of context length and computational efficiency. Existing models are constrained by relatively short context windows and lack modern optimizations, which are adopted in more recent architectures like ModernBERT. To address this gap, ModBERTBr~\cite{Ben2025} was proposed as an initial application of the ModernBERT architecture to Portuguese. Unlike BERTimbau, ModBERTBr was trained from scratch with randomly initialized weights using Portuguese Wikipedia~\cite{Wikipedia} and the BrWaC corpus. However, this approach only partially realized the architectural potential of ModernBERT, as it relied on a relatively small dataset and failed to explore context expansion, ultimately remaining limited to the legacy 512-token window of its predecessors.

\section{BERTomelo}
\label{sec:bertomelo}

BERTomelo is a Portuguese BERT encoder built upon the ModernBERT~\cite{Warner2024} architecture and trained from scratch on the ClassiCC-PT corpus~\cite{Almeida2025}. BERTomelo is available in two versions, Base and Large, both adopting a 1,024-token context window. The model incorporates efficiency mechanisms inherent to the ModernBERT architecture, such as FlashAttention~\cite{Dao2022}, Unpadding~\cite{Zeng2022}, and Alternating Attention. The model achieves state-of-the-art performance on NER tasks and competitive results on other benchmarks.

Regarding the data, existing Portuguese BERT models primarily rely on the BrWaC corpus~\cite{Filho2018}, which comprises 3.53 million documents and was released in 2018. In the context of BERTomelo's development, the BrWaC corpus was considered significantly outdated and insufficient to leverage the high-throughput processing capabilities of the ModernBERT~\cite{Warner2024} architecture. Given the availability of more contemporary corpora, ClassiCC-PT\footnote{\url{https://huggingface.co/datasets/ClassiCC-Corpus/ClassiCC-PT}}~\cite{Almeida2025} was selected as the training source. This large-scale web corpus contains 96 million Portuguese documents extracted from Common Crawl snapshots, processed with a focus on data quality, language specificity, deduplication, and targeted filtering.

The adopted tokenizer should be developed with the same monolingual scope as the training corpus~\cite{Rust2020}. Following this principle, BERTomelo employs the same tokenizer utilized by ModBERTBr~\cite{Ben2025}, a custom fast tokenizer based on the Unigram~\cite{Kudo2018} algorithm. This approach was selected due to related literature showing that Unigram models demonstrate superior performance when applied to Romance languages such as Spanish and French~\cite{Ali2024}. The tokenizer was trained on a processed subset of a combined corpus consisting of Wikipedia~\cite{Wikipedia} and BrWaC~\cite{Filho2018}, with a vocabulary size of 32,768 tokens.

Although there is a discrepancy between the original tokenizer's training corpus and the ClassiCC-PT used for pre-training, it is assumed that the combination of BrWaC and Wikipedia provides a robust lexical baseline for Portuguese. Reusing this component significantly reduces the computational overhead associated with training a new tokenizer from scratch. Moreover, the tokenization pipeline remains identical to the one utilized by ModBERTBr, beginning with a normalization stage (lowercase conversion, NFC Unicode normalization, and quote standardization). This is followed by a metaspace-based pre-tokenization. Finally, the post-processing stage appends standard \texttt{[CLS]} and \texttt{[SEP]} tokens to define sequence boundaries and maintain compatibility with legacy BERT architectures.

Table \ref{tab:architecture} presents the architectural parameters of BERTomelo. Both versions are built upon their respective ModernBERT~\cite{Warner2024} architectures and adopt a 1,024-token context window. The Base version features 22 layers, a hidden size of 768, and 12 attention heads, while the Large version consists of 28 layers, a hidden size of 1,024, and 16 attention heads. To support open-source research and community access, both models have been made publicly available on the Hugging Face Hub\footnote{\url{https://huggingface.co/unb-labia/BERTomelo-ModernBERT-Base-v1}, \url{https://huggingface.co/unb-labia/BERTomelo-ModernBERT-Large-v1}} with the source code hosted on GitHub\footnote{\url{https://github.com/unb-labia/bertomelo-brazilian-modernbert}}. A primary objective of BERTomelo is the systematic integration of ModernBERT’s architectural and efficiency advancements into the pre-training process for Brazilian Portuguese, ensuring technical alignment throughout its development. Specifically, the architecture incorporates bias-free linear layers,  pre-normalization blocks, Gaussian Error Gated Linear Unit (GeGLU) activation, and RoPE~\cite{Su2023}. Furthermore, efficiency enhancements such as Alternating Attention, Unpadding~\cite{Zeng2022}, and FlashAttention-2~\cite{Dao2023} significantly improve token and information throughput.

\begin{table}[htb]
\centering
\caption{Architectural specifications for BERTomelo Base and Large.}
\label{tab:architecture}
\begin{tabular}{lll}
\toprule
\textbf{Parameter} & \textbf{Base} & \textbf{Large} \\
\midrule
\textit{Parameters} & 136.120.832 & 377.841.664\\
\textit{Activation function} & GeGLU & GeGLU \\
\textit{Hidden size} & 768 & 1,024 \\
\textit{Intermediate size} & 1,152 & 1,536 \\
\textit{Attention heads} & 12 & 16 \\
\textit{Number of hidden layers} & 22 & 28 \\
\textit{Local attention size} & 128 & 128 \\
\textit{Context window} & 1,024 & 1,024 \\
\textit{Vocabulary size} & 32,768 & 32,768 \\
\bottomrule
\end{tabular}
\end{table}

\section{Methodology}
\label{sec:experiments}

This section describes the methodology and parameters selected for the pre-training of BERTomelo, as well as the evaluation tasks and metrics used for performance comparison. Section \ref{subsec:preprocessing} discusses the preprocessing strategy for splitting the corpus. Section \ref{subsec:training} presents the training protocol adopted for both the Base and Large versions, and Section \ref{subsec:evaluation} discusses the evaluation procedures of the trained models.

\subsection{Preprocessing}
\label{subsec:preprocessing}

The corpus is split at an early stage of the pipeline to ensure that each set can be independently preprocessed and manipulated without the risk of cross-contamination. The resulting distribution consists of a training set containing 80\% of the documents, with the evaluation and test sets each comprising 10\%. For the training set, the first preprocessing step involves segmenting each document using the newline character, dividing the corpus into smaller sequences that better align with the proposed 1,024-token context window. This segmentation is applied exclusively to the training set, while the validation and test sets remain unaltered in order to better reflect raw, real-world input data.

Following segmentation, quality filtering criteria are applied based on those proposed for the Gopher model~\cite{Rae2022} and later adopted in ModBERTBr~\cite{Ben2025}. Each sequence is evaluated according to its mean word length, retaining only those within the range of 2 to 15 characters, which helps remove texts with incorrectly spaced or improperly concatenated words. Additionally, a minimum of 2 stop words per sequence is required to preserve syntactic and semantic coherence. While ModBERTBr employs a threshold of a single stop word, a stricter requirement is adopted in this work to favor longer and more informative sequences that align with BERTomelo's extended context window. Subsequently, sequences are constrained by length, retaining only those containing between 20 and 1,024 words. As a result, some sequences may exceed the 1,024-token limit after tokenization, in which case they are truncated.

% \begin{table}[htb]
% \centering
% \caption{Preprocessing parameters.}
% \label{tab:preprocessing}
% \begin{tabular}{ll}
% \toprule
% \textbf{Parameter} & \textbf{Value} \\
% \midrule
% \textit{Segmentation method} & Split by \texttt{"\textbackslash n"} \\
% \textit{Min. Number of Words} & 20 \\
% \textit{Max. Number of Words} & 1,024 \\
% \textit{Min. Number of Stopwords} & 2 \\
% \textit{Min. Average Size of Words} & 2 \\
% \textit{Max. Average Size of Words} & 15 \\
% \bottomrule
% \end{tabular}
% \end{table}

\subsection{Training Protocol}
\label{subsec:training}

The training protocol of BERTomelo encompasses both Base and Large versions. The strategy is closely aligned with the ModernBERT~\cite{Warner2024} approach to ensure consistency between the architecture and the training strategy. The models are initialized from scratch with random weights. The training is conducted within the Hugging Face ecosystem. Following ModernBERT~\cite{Warner2024}, the NSP task is omitted, leaving MLM as the sole training objective. An increased masking rate of 30\% is adopted.

Regarding attention mechanisms, the Alternating Attention strategy is employed with a local attention window size of 128 tokens. Specifically, two consecutive layers apply local attention, followed by a layer with global attention. For positional encoding, the base frequency $\theta$ of RoPE is adjusted from the default value of 160,000 to 10,000. While the original value is designed for an 8,192-token context window, this adjustment is better suited for the current 1,024-token setup and facilitates future context scaling through continued pre-training~\cite{Su2023}.

Both models were trained using an AMD High-Performance Computing cluster. The Base model was trained on 8 AMD Instinct MI300X GPUs, while the Large model utilized 8 AMD Instinct MI325X GPUs. Both models share the same training configuration, with a micro-batch size of 256 per GPU and no gradient accumulation, resulting in a global batch size of 2048. This setup achieves stable GPU utilization, typically ranging between 50\% and 60\% of VRAM usage. Although larger batch sizes were explored, they triggered instabilities, such as weight divergence and out-of-memory errors. To further ensure stability, training was performed using \texttt{bf16} mixed precision. Table \ref{tab:training_unified} summarizes the main hyperparameters and specifications of this setup.

\begin{table}[htb]
\centering
\caption{Specifications of the BERTomelo training Protocol.}
\label{tab:training_unified}
\begin{tabular}{lcc}
\toprule
\textbf{Parameter} & \textbf{Base} & \textbf{Large} \\
\midrule
\textit{Max training steps} & \multicolumn{2}{c}{2,000,000} \\
\textit{Micro batch size} & \multicolumn{2}{c}{256} \\
\textit{Global batch size} & \multicolumn{2}{c}{2,048} \\
\textit{Optimizer} & \multicolumn{2}{c}{AdamW} \\
\textit{Learning rate} & \multicolumn{2}{c}{$5 \times 10^{-5}$} \\
\textit{Weight decay} & \multicolumn{2}{c}{0.0} \\
\textit{Learning rate scheduler} & \multicolumn{2}{c}{Linear} \\
% \textit{Evaluation frequency} & \multicolumn{2}{c}{10,000 steps} \\
% \textit{Early stop patience} & \multicolumn{2}{c}{4} \\
% \textit{RoPE $\theta$} & \multicolumn{2}{c}{10,000} \\
% \textit{Precision} & \multicolumn{2}{c}{bf16 mixed precision} \\
% \midrule
% \textit{Hardware} & 8x AMD Instinct MI300X & 8x AMD Instinct MI325X \\
% \textit{Training duration} & $\sim$121 hours & $\sim$170 hours \\
% \bottomrule
\end{tabular}
\end{table}

The number of training steps is set to a maximum of 2 million steps, with evaluation performed every 10,000 steps. During evaluation, the model is assessed on a fixed shuffled subset of 1.5 million sequences from the validation set. The evaluation loss is used both for early stopping and for selecting the best model checkpoint. Although early stopping is configured with a patience of four evaluation cycles, neither model triggered this criterion, suggesting that further training or a tighter stopping threshold could be explored. %Furthermore, even if the final evaluation shows a slight increase, it does not necessarily guarantee that the model has reached full saturation. 
The total training time is approximately 121 hours for the Base model and 170 hours for the Large model. During this process, approximately 386.48 billion tokens are processed, comprising 8.67 epochs for each model version.

\subsection{Evaluation Strategy}
\label{subsec:evaluation}

For the evaluation of BERTomelo, the strategy established by ModBERTBr~\cite{Ben2025} is followed. The pre-trained models are subjected to intrinsic and extrinsic evaluations. In the intrinsic evaluation, the tokenizer is assessed using the fertility metric, and the proportion of unknown tokens is computed. Training stability is monitored through an analysis of the loss and gradient norm. Finally, a primary performance metric is provided by the evaluation loss, calculated by applying the MLM task on a representative portion of the test set covering 3.6 million sequences ($37.5\%$ of the total).

Subsequently, the extrinsic analysis consists of fine-tuning both pre-trained model variants on downstream tasks. The objective is to evaluate the overall performance of the model in NLP applications and establish a comparative baseline with related work. To ensure backward compatibility, the benchmarks utilized for the evaluation of ModBERTBr are maintained. These encompass both sentence and token-level analysis across the STS, RTE, and NER tasks.

For STS and RTE tasks, the ASSIN2\footnote{\url{https://huggingface.co/datasets/nilc-nlp/assin2}}~\cite{Real2020} corpus is used. It consists of manually annotated sentence pairs in Brazilian Portuguese. For the STS task, the extracted metrics include the Mean Squared Error (MSE), which measures the difference between the predicted and annotated labels, and the Pearson correlation coefficient, which reflects how the model's similarity scores align with the annotated scores. For RTE, accuracy and Macro F1-score are used to evaluate entailment recognition performance. Finally, the evaluation of the NER task is grounded in the LeNER-Br\footnote{\url{https://huggingface.co/datasets/peluz/lener\_br}}~\cite{Luz2018} dataset, which consists of Brazilian legal documents. The performance is assessed via precision, recall, and Micro F1-score, standard metrics for assessing named entity classification.

To ensure reproducibility, the fine-tuning process adheres to specific hyperparameter configurations for each task. These settings --- including maximum training steps, batch size, learning rate (LR), and scheduling policies --- are detailed in Table \ref{tab:finetuning_evaluation}. Specifically, the NER and STS tasks employ an LR with weight decay and a Warmup-Stable-Decay (WSD) scheduler, while the RTE task follows a linear decay schedule.

\begin{table}[!htb]
\centering
\caption{Finetuning hyperparameters for extrinsic evaluation tasks.}
\label{tab:finetuning_evaluation}

    \begin{tabular}{lccc}
    \toprule
    \textbf{Hyperparameter} & \textbf{STS (ASSIN2)} & \textbf{RTE (ASSIN2)} & \textbf{NER (LeNER-Br)} \\
    \midrule
    \textit{Max steps} & 5,000 & 10,000 & 3,000 \\
    \textit{Batch size} (Train/Eval) & 4 / 4 & 4 / 4 & 4 / 32 \\
    \textit{Learning rate} & $2 \times 10^{-5}$ & $1 \times 10^{-5}$ & $2 \times 10^{-5}$ \\
    \textit{Weight Decay} & $1 \times 10^{-4}$ & $1 \times 10^{-4}$ & $1 \times 10^{-4}$ \\
    \textit{Learning rate scheduler} & WSD & Linear & WSD \\ 
    \bottomrule
    \end{tabular}

\end{table}

Baseline results are compared with the performance reported by Wu et al. ~\cite{Ben2025}, adhering to their fine-tuning protocols. These comparisons include the Base variants of ModernBERT~\cite{Warner2024}, BERTimbau~\cite{Souza2020}, BERTuguês~\cite{MazzaZago2024}, mBERT~\cite{Devlin2018}, and ModBERTBr~\cite{Ben2025}. The selected Large variants are BERTimbau Large and Albertina~\cite{Rodrigues2023} 900M for Portuguese models, while XLM-RoBERTa Large is used as a representative multilingual model.

\section{Experiments and Results}
\label{sec:results}

This section details the results of the training and evaluation protocols. It is structured as follows: Section \ref{sec:intrinsic} presents the intrinsic evaluation, including training metrics, tokenizer performance, and stability monitoring. Section \ref{sec:extrinsic} discusses the extrinsic evaluation results obtained across various downstream tasks. 

\subsection{Intrinsic Evaluation}
\label{sec:intrinsic}

The suitability of the tokenizer adopted from ModBERTBr is verified through the fertility metric and the unknown token ratio calculated on the training set. This assessment evaluates whether the vocabulary effectively represents contemporary textual sequences. The fertility metric, which indicates the average number of tokens generated per word, measures subword fragmentation efficiency. The results show a fertility of $1.506$ and an unknown token ratio of 0.0017\%. These values corroborate the assumption that the tokenizer is well-suited for the ClassiCC-PT~\cite{Almeida2025} corpus. 

\begin{figure*}[htp]
\label{fig:large_graph}
  \centering
  \subfigure[Base Loss]{\includegraphics[scale=0.3]{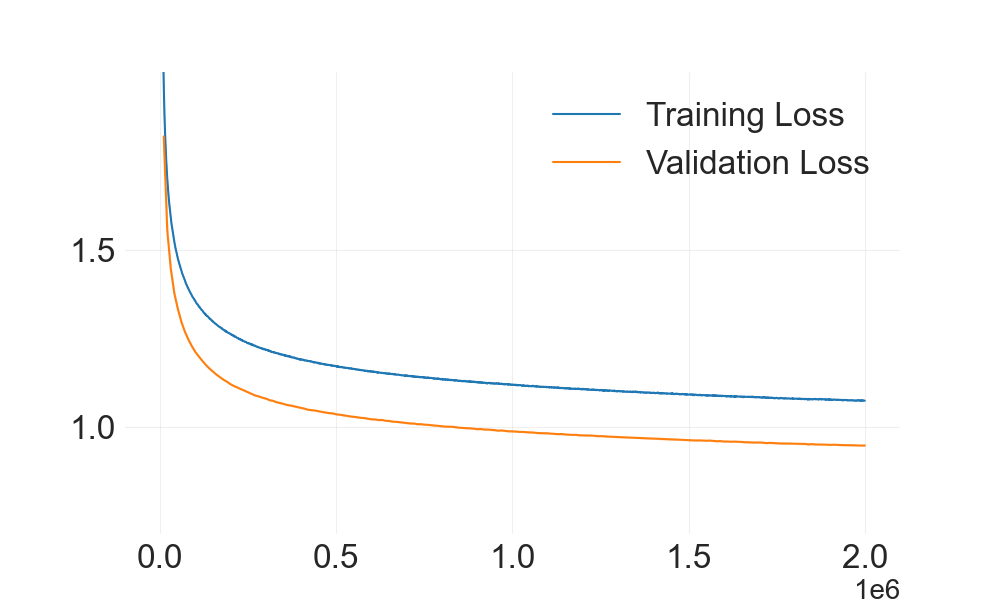}\label{fig:loss_base}}
  \subfigure[Large Loss]{\includegraphics[scale=0.3]{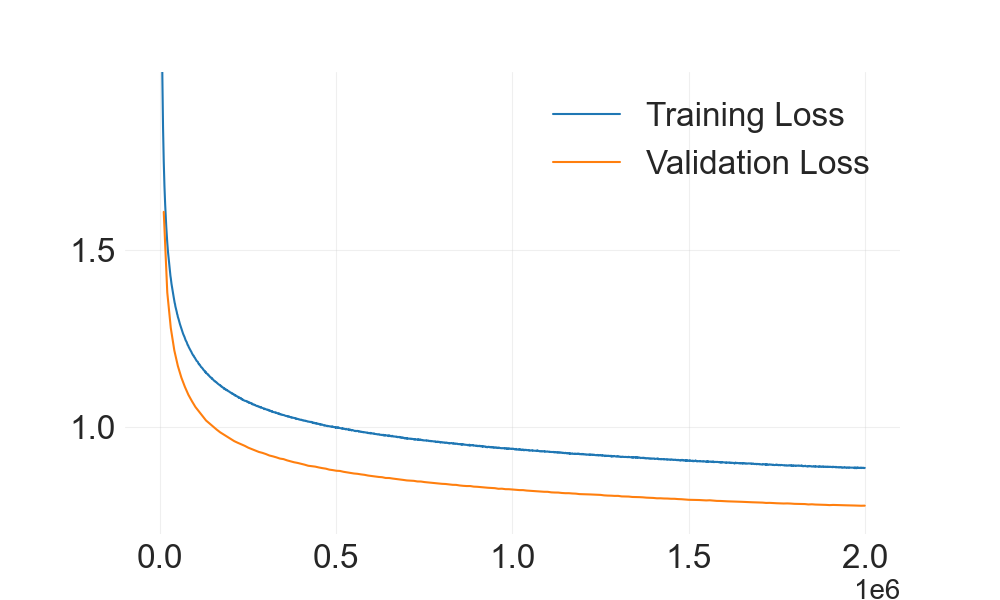}\label{fig:large_loss}}\quad
  \subfigure[Base Gradient Norm]{\includegraphics[scale=0.3]{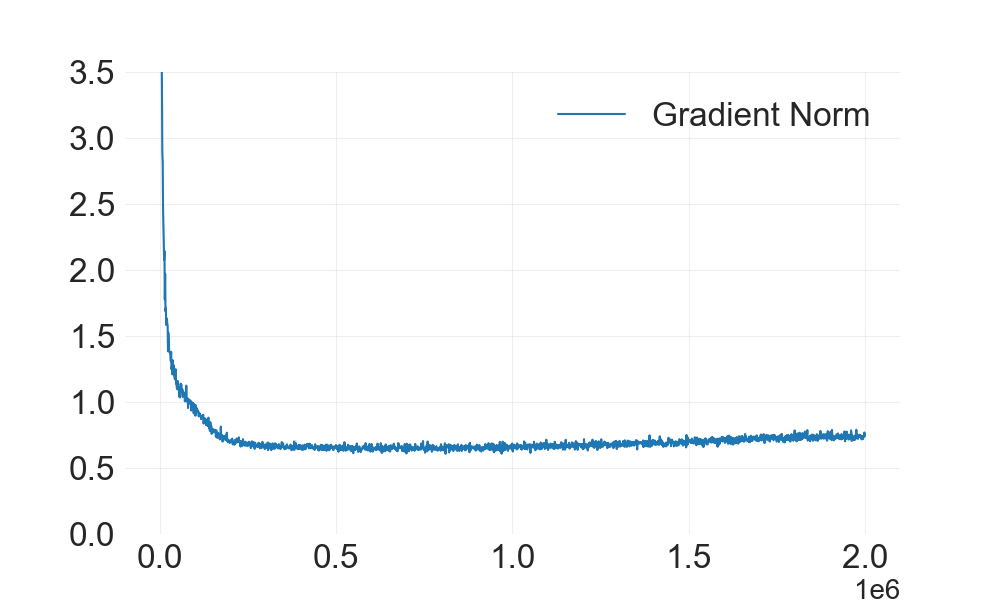}\label{fig:grad_base}}
  \subfigure[Large Gradient Norm]{\includegraphics[scale=0.3]{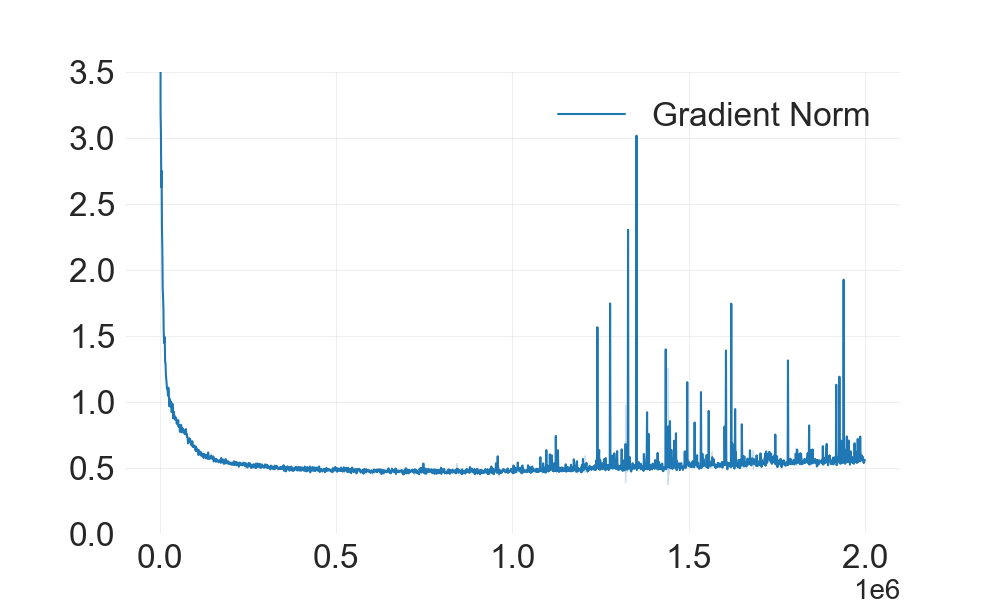}\label{fig:grad_large}}
  \caption{Training metrics for the BERTomelo variants.}
  \label{fig:training_metrics}
\end{figure*}

Training stability is monitored through loss and gradient norm metrics, with results for both the Base and Large versions consolidated in Figure \ref{fig:training_metrics}. In these plots, the $x$-axis represents the training steps, while the $y$-axis denotes the respective metric values. The training metrics are recorded every 1,000 steps, whereas evaluation metrics are computed every 10,000 steps. The plots are presented individually as follows: Figures \ref{fig:loss_base} and \ref{fig:large_loss} display the training and evaluation loss, while Figures \ref{fig:grad_base} and \ref{fig:grad_large} show the gradient norm for the Base and Large variants, respectively.

The training results indicate stable convergence for both BERTomelo versions. Although the process did not trigger early stopping criteria, the loss curves exhibit a smooth, progressive decay toward saturation. Notably, the evaluation loss remains consistently lower than the training loss, which can be attributed to the absence of regularization mechanisms, such as dropout, during evaluation. The aligned decay between both metrics further demonstrates the model's generalization capability on unseen data. Moreover, while the Base model maintains a consistent gradient norm, the Large model presents transient spikes after the 1 million step mark. However, these fluctuations do not compromise overall stability, as the norm consistently returns to its baseline scale.

Following pre-training, the models are evaluated on the MLM task using a portion of $37.5\%$ of the test set, covering 3.6 million sequences. The Base version achieves an accuracy of $79.26\%$ with a test loss of $0.9495$, while the Large version reaches $82.28\%$ and $0.7799$, respectively. These results align with the observed loss decay during training, indicating a robust capture of the Portuguese syntactic and semantic patterns. Notably, the Large model consistently outperforms the Base model, leveraging its increased representative capacity to achieve superior performance. Overall, the results confirm that the BERTomelo models generalize effectively to unseen data and provide a solid backbone for subsequent downstream tasks.

\subsection{Extrinsic Evaluation}
\label{sec:extrinsic}

% The extrinsic evaluation comprises the results of metrics obtained after fine tuning BERTomelo on downstream tasks and evaluates its performance on the test set of each benchmark corpus. Baseline results for the compared models are reported by Wu et al. (2025). The authors perform fine tuning on the related models, with the exception of BERTimbau, for which the metrics are sourced from the original study. These results are detailed in Table \ref{tab:model_comparison}, which displays both variants results compared to models of similar scale.

Table \ref{tab:model_comparison} presents the extrinsic evaluation results, categorized into Base and Large model versions. The performance is assessed across three downstream benchmarks: STS and RTE, both using the ASSIN2 corpus, and NER via the LeNER-Br corpus. In this comparison, bold values denote the best performance within each scale category. Baseline results for the compared models are sourced from Wu et al. ~\cite{Ben2025}, who provided fine-tuning metrics for most related models, while the results for BERTimbau are reported from its original study. This setup enables a direct comparison of both BERTomelo variants against state-of-the-art models of equivalent scale under consistent test conditions.

\begin{table}[ht]
\centering
\renewcommand{\arraystretch}{1.4}
\caption{Models comparison results on downstream tasks (best in bold). Large models were fine-tuned with the same hyperparameters.}
\label{tab:model_comparison}
\resizebox{\textwidth}{!}{%
\begin{tabular}{@{}lcccccccc@{}}
\toprule
\multirow{2}{*}{\textbf{Model}} & \multicolumn{2}{c}{\textbf{STS (ASSIN 2)}} & \multicolumn{2}{c}{\textbf{RTE (ASSIN 2)}} & \multicolumn{3}{c}{\textbf{NER (LeNER-Br)}} \\ \cmidrule(lr){2-3} \cmidrule(lr){4-5} \cmidrule(lr){6-8}
 & MSE $\downarrow$ & Pearson $\uparrow$ & Macro F1-Score $\uparrow$ & Acc $\uparrow$ & Micro F1-Score $\uparrow$ & Recall $\uparrow$ & Prec $\uparrow$ \\ \midrule
mBERT$^*$ & 0.597 & 0.801 & 84.45\% & 84.52\% & 88.50\% & 90.45\% & 86.63\% \\
BERTimbau$^*$ & 0.580 & \textbf{0.836} & \textbf{89.20\%} & \textbf{89.20\%} & 90.48\% & 91.64\% & 89.35\% \\
BERTuguês$^*$ & 0.583 & 0.823 & 86.27\% & 86.40\% & 89.56\% & 90.59\% & 88.55\% \\
ModernBERT$^*$ & 0.514 & 0.790 & 81.09\% & 81.17\% & 75.76\% & 78.03\% & 73.61\% \\
ModBERTBr$^*$ & 0.509 & 0.812 & 85.28\% & 85.42\% & 90.08\% & \textbf{92.40\%} & 87.88\% \\ \midrule
\textbf{BERTomelo Base} & \textbf{0.425} & 0.833 & 87.65\% & 87.70\% & \textbf{91.37\%} & 92.29\% & \textbf{90.46\%} \\ \midrule
BERTimbau Large & 0.500 & 0.852 & \textbf{90.04\%} & \textbf{90.04\%} & 90.54\% & 92.24\% & 88.90\% \\
Albertina-900M Pt-Br & 0.570 & \textbf{0.853} & 89.09\% & 89.09\% & 87.28\% & 91.43\% & 83.49\% \\
XLM Roberta Large & 0.609 & 0.804 & 72.53\% & 73.32\% & 87.40\% & 90.13\% & 84.84\% \\ \midrule
\textbf{BERTomelo Large} & \textbf{0.401} & 0.849 & 89.21\% & 89.26\% & \textbf{91.05\%} & \textbf{92.75}\% & \textbf{89.42\%} \\ \bottomrule
\multicolumn{8}{l}{\footnotesize $^*$ Baseline results retrieved from \cite{Ben2025}. BERTimbau Large results on ASSIN 2 are retrieved from \cite{Souza2020}.} \\
\end{tabular}%
}
\end{table}

In the proposed evaluation tasks, BERTomelo achieves the lowest MSE in the STS task for both variants, demonstrating that the model predicts similarity scores with a high fidelity to the ground truth. The marginal difference of 0.003 in the Pearson coefficient between the BERTomelo Base variant and BERTimbau reflects a comparable linear relationship in their scores. Regarding RTE, BERTomelo's performance is surpassed only by BERTimbau. In the NER task, BERTomelo achieves the highest Micro F1-score and Precision for the Base variant, while also securing the highest Recall for the Large version.

A notable observation derived from the results is the performance advantage of Portuguese-specific models relative to multilingual alternatives. This disparity validates the decision to pursue a dedicated monolingual pre-training approach.  Although the 1024-token context window of BERTomelo is not explicitly evaluated within the current benchmarks, this characteristic merits consideration when interpreting the results, as it represents a core capability for processing longer-range dependencies.

\section{Conclusion and Future Works}
\label{sec:conclusion}

BERTomelo is proposed as a suite of encoder-only models designed to bridge the architectural gap in the Brazilian Portuguese NLP landscape, which has long relied on legacy BERT frameworks. By integrating the state-of-the-art innovations of ModernBERT, BERTomelo delivers a robust, tailored pre-training process that features an extended context window of 1,024 tokens and significant efficiency gains. The suite comprises the Base and Large variants. Both architectures implement critical advancements, including the removal of bias terms, pre-normalization blocks, GeGLU activation, and RoPE for positional encoding. Furthermore, the integration of FlashAttention-2, Alternating Attention, and Unpadding mechanisms ensures high token throughput and training stability. Experimental results demonstrate that BERTomelo achieves highly competitive performance relative to existing models, effectively validating the systematic integration of these architectural improvements for the Portuguese language.

The evaluation of both versions yielded robust performance across both intrinsic and extrinsic evaluation strategies. The intrinsic analysis validated the effectiveness of the training protocol and the selection of a tokenizer tailored to the training corpus. Extrinsically, BERTomelo achieved competitive results, notably outperforming existing models in NER and STS tasks while maintaining robust metrics in RTE. Ultimately, these findings reinforce that models specifically optimized for Brazilian Portuguese demonstrate superior performance in monolingual benchmarks compared to multilingual alternatives, justifying the necessity of dedicated architectural adaptation.

Future work involves leveraging the RoPE implementation already integrated into the BERTomelo architecture to extend the context window. While the current version provides a 1,024 token capacity, twice the context window of standard baseline models, the next phase aims to expand this limit to 8,192 tokens, fully utilizing ModernBERT's inherent architectural support for long-range dependencies. Additionally, the implementation of sequence packing is planned to maximize token throughput and optimize token processing efficiency during training. The evaluation framework will also be expanded to include benchmarks specifically designed for long document analysis, ensuring a rigorous assessment of the model's performance on extended contexts. Finally, further investigation into preprocessing and segmentation protocols will be conducted to determine the optimal strategy for handling significantly larger sequence lengths while maintaining competitive performance across diverse downstream tasks.

\begingroup
\small

\vspace{12pt}\noindent
\textbf{Acknowledgments}\hspace{0.5em} This work was supported in part by Advanced Micro Devices, Inc. under the AMD AI \& HPC Cluster Program. Furthermore, the respective authors are appreciated for providing the ClassiCC-PT, ASSIN2, and LeNER-BR datasets.

\endgroup

\bibliographystyle{unsrt}  
\bibliography{references} 

\end{document}